\documentclass[letterpaper]{article} % DO NOT CHANGE THIS
\usepackage{aaai24}  % DO NOT CHANGE THIS
\usepackage{times}  % DO NOT CHANGE THIS
\usepackage{helvet}  % DO NOT CHANGE THIS
\usepackage{courier}  % DO NOT CHANGE THIS
\usepackage[hyphens]{url}  % DO NOT CHANGE THIS
\usepackage{graphicx} % DO NOT CHANGE THIS
\usepackage{natbib}  % DO NOT CHANGE THIS AND DO NOT ADD ANY OPTIONS TO IT
\usepackage{caption} % DO NOT CHANGE THIS AND DO NOT ADD ANY OPTIONS TO IT
\usepackage{algorithm}
\usepackage{algorithmic}
\usepackage{amssymb}
\usepackage{microtype}
\usepackage{multirow}
\usepackage{booktabs}
\usepackage{float}
\usepackage{amsmath}
\usepackage{lipsum}
\usepackage{newfloat}
\usepackage{listings}
\DeclareCaptionStyle{ruled}{labelfont=normalfont,labelsep=colon,strut=off} % DO NOT CHANGE THIS
\floatstyle{ruled}
\newfloat{listing}{tb}{lst}{}
\floatname{listing}{Listing}
\title{CPDM: Content-Preserving Diffusion Model for Underwater Image Enhancement}
\title{CPDM: Content-Preserving Diffusion Model for Underwater Image Enhancement}
\author {
    Xiaowen Shi and
    Yuan-Gen Wang
}
\affiliations {
    School of Computer Science and Cyber Engineering, Guangzhou University, China\\
    shixiaowen@e.gzhu.edu.cn, wangyg@gzhu.edu.cn
}
\begin{document}

\maketitle
%----------Fig1.-----------
%--------------------------
\begin{figure*}[th]
% \begin{minipage}[b]{1.0\linewidth}
  \centering
  % \centerline{\includegraphics[width=8.5cm]{fig1.png}}
  \centerline{\includegraphics[width=18cm]{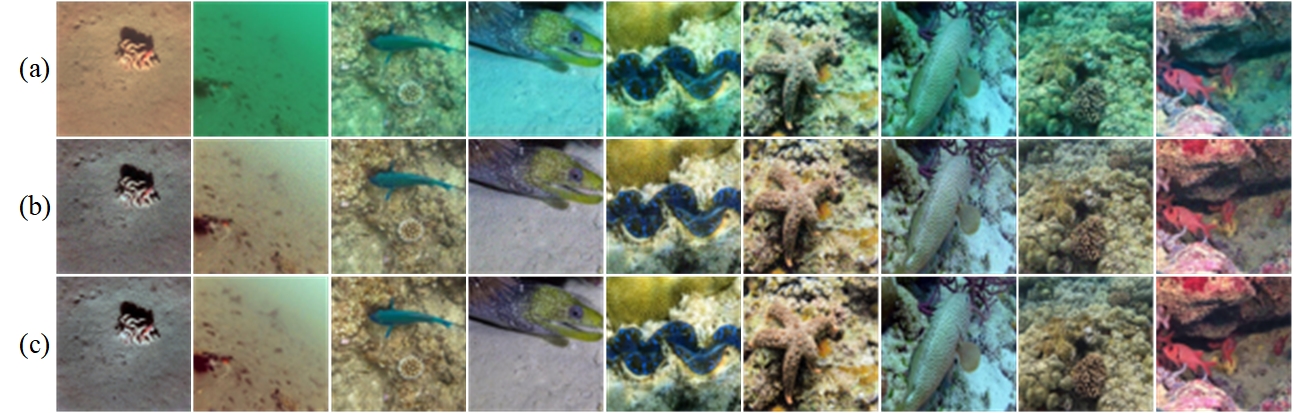}}

  % \centerline{(a) Result 1}\medskip
% \end{minipage}
\vspace{-0.5em}
\caption{A visual illustration of the proposed CPDM method. (a) Original images to be enhanced. (b) Images enhanced by our CPDM. (c) Reference images as ground truth.}
\label{fig:res}
 % \vspace{-1.4em}
\end{figure*}

\begin{abstract}
Underwater image enhancement (UIE) is challenging since image degradation in aquatic environments is complicated and changing over time. Existing mainstream methods rely on either physical-model or data-driven, suffering from performance bottlenecks due to changes in imaging conditions or training instability. In this article, we make the first attempt to adapt the diffusion model to the UIE task and propose a Content-Preserving Diffusion Model (CPDM) to address the above challenges. CPDM first leverages a diffusion model as its fundamental model for stable training and then designs a content-preserving framework to deal with changes in imaging conditions. Specifically, we construct a conditional input module by adopting both the raw image and the difference between the raw and noisy images as the input,  which can enhance the model's adaptability by considering the changes involving the raw images in underwater environments. To preserve the essential content of the raw images, we construct a content compensation module for content-aware training by extracting low-level features from the raw images. Extensive experimental results validate the effectiveness of our CPDM, surpassing the state-of-the-art methods in terms of both subjective and objective metrics. 
%The code is available at https://github.com/GZHU-DVL/CPDM.
\end{abstract}

%the raw underwater images as conditions for model training. Additionally, the difference between the raw underwater images and the noisy images at each step is adopted as supplementary training conditions. 

\section{Introduction}
Underwater image enhancement (UIE) has gained great attention recently as human activities have increasingly ventured into the ocean. However, enhancing the quality of degraded underwater images poses significant challenges due to the complex and ever-changing underwater environment, as well as poor lighting conditions. The degradation of underwater images is primarily caused by the selective absorption and scattering of visible light wavelengths within the underwater environment \cite{mcglamery1980computer, jaffe1990computer, hou2012optical,akkaynak2017space,akkaynak2018revised}. Consequently, the acquired underwater images exhibit low contrast, low brightness, significant color deviations, blurred details, uneven bright spots, and other defects. These limitations greatly hinder practical applications in fields such as marine ecology \cite{strachan1993recognition}, marine biology and archaeology \cite{ludvigsen2007applications}, remotely operated vehicles, and autonomous underwater vehicles \cite{johnsen2016use,ahn2017enhancement}. Therefore, the study of UIE holds immense significance for advancing related underwater research.

Some UIE techniques have been developed for enhancing the quality of underwater images, which can be roughly categorized into physical-model and data-driven methods. Physical-model methods \cite{galdran2015automatic,li2016underwater1, drews2016underwater,li2017hybrid,peng2017underwater, wang2017single,peng2018generalization,akkaynak2019sea} aim to model the physical process of light propagation in water by taking absorption, scattering, and other optical properties of the underwater environment into account. These methods often involve complex mathematical models and algorithms to simulate degradation. However, since the aquatic environments change over time, the method established in a certain physical scenario cannot adapt to other different physical scenarios, resulting in poor generalization. 

Motivated by the success of deep learning in a wide range of fields, data-driven methods \cite{li2017watergan, li2018emerging, guo2019underwater, 9426457ucolor, fu2022underwater, peng2023u} have been proposed by learning the mapping between degraded underwater images and their corresponding high-quality reference images. These methods rely on large-scale datasets for model training, effectively enhancing image quality based on learned patterns and features. However, currently established UIE datasets are generally collected in a specific underwater environment, such as low lighting, various turbidity, and different densities of particulate matter. Hence, the model trained on a single dataset suffers from poor cross-dataset performance.

In this work, we propose a novel UIE framework, termed Content-Preserving Diffusion Model (CPDM), for enhancing the quality of underwater images. Specifically, we utilize the raw image as a conditional input during the model training process. To facilitate the extraction of differential features at different time steps, we introduce the differences between the raw image and the noisy image at each time step as another conditional input. Furthermore, to ensure that the model preserves the essential content of raw images, we design a content compensation module to extract the low-level features of raw images for content-aware training. Figure \ref{fig:res} provides a preview of the enhancement results achieved by our CPDM. The main contributions of this article are as follows:

\begin{itemize}
\item We present a Content-Preserving Diffusion Model (CPDM) for underwater image enhancement (UIE). To adapt the diffusion model to such a new task, we take the raw image as conditional input for training at each time step, enabling the restored target image to have consistent content with the raw image. 

\item We introduce the difference between the raw image and the noisy image of the current time step as an extra conditional input for content-aware training at the current time step, iteratively refining the output of each time step and resulting in a high-quality enhanced target image.

\item We design a content compensation module to ensure that the trained model preserves the low-level features of raw images, such as structure, texture, and edge, preventing excessive modification of these important details within the images.

\item Extensive experimental results show that our CPDM outperforms state-of-the-art methods. The ablation study also demonstrates the effectiveness of each module designed in this work. Especially, the proposed CPDM achieves good generalization and greatly improves color fidelity.
\end{itemize}

\section{Related Work}

\subsection{Underwater Image Enhancement Methods}
Methods for underwater image enhancement play a crucial role in improving the visual quality of underwater images. The existing mainstream methods mainly consist of physical-model and data-driven \cite{li2019underwater}.

\noindent \textbf{Physical-model.} Physical-model methods treat underwater image enhancement as an inverse problem. These methods involve several steps: constructing a physical model to simulate the degradation process, estimating unknown model parameters based on the given images, and solving the inverse problem using the estimated parameters. By solving the inverse problem, these methods can enhance underwater image quality, mitigating the impact of degradation factors such as light attenuation, scattering, and color distortion. Drews et al. \cite{drews2016underwater} introduced the Underwater Dark Channel Prior method, addressing the issue of the unreliable red channel in underwater images. Liu et al. \cite{liu2016underwater} formulated a cost function based on the observation that the dark channel of underwater images tends to zero and minimized it to find the optimal transmission mapping that maximizes image contrast. Peng et al. \cite{peng2017underwater} proposed a method for enhancing underwater images by estimating image blur and depth. Peng et al. \cite{peng2018generalization} introduced the Generalized Dark Channel Prior to image restoration, which incorporates adaptive color correction into the image formation model. Akkaynak et al. \cite{akkaynak2019sea} presented a modified underwater color correction method for enhancing underwater images.

\noindent \textbf{Data-driven.} In comparison to physical-model methods, data-driven methods for underwater image enhancement have been developed lately \cite{2013Transmission, 2019Underwater}. Since the effectiveness of underwater image enhancement is influenced by specific factors such as scene, lighting condition, temperature, and turbidity, it is challenging to employ synthetic and realistic underwater images for network training. Moreover, neural networks trained on synthetic underwater images may not generalize well to real-world scenarios. WaterGAN \cite{li2017watergan} utilized in-air images and depth maps as input to generate synthetic underwater images as output. These synthetic underwater images are then used for color correction of monocular underwater images. Water CycleGAN \cite{li2018emerging} relaxed the requirement of paired underwater images by utilizing a weakly-supervised color transfer approach to correct color distortions. However, it may generate unrealistic results. Guo et al. \cite{guo2019underwater} proposed a multi-scale dense GAN for underwater image enhancement. However, this method still cannot overcome the limitations of unpredicted outputs from GANs \cite{goodfellow2014gan}. Ucolor \cite{9426457ucolor} integrated an underwater physical imaging model and a medium transmission-guided model to enhance image quality in regions with severe degradation. However, the performance of the approach is significantly affected by different underwater environments. U-shape \cite{peng2023u} proposed a U-shape Transformer with integrated modules to reinforce the network's attention to color channels and spatial areas that suffer from severe attenuation. This model includes a channel-wise multi-scale feature fusion transformer module and a spatial-wise global feature modeling transformer module. However, this method still exhibits major color distortion.

%----------Fig2.-----------
%--------------------------
\begin{figure*}[htbp]
% \begin{minipage}[b]{1.0\linewidth}
  \centering
  % \centerline{\includegraphics[width=8.5cm]{fig1.png}}
  % \centerline{\includegraphics[height=5cm, width=13.5cm]{LaTeX/fig2.png}}
  \centerline{\includegraphics[]{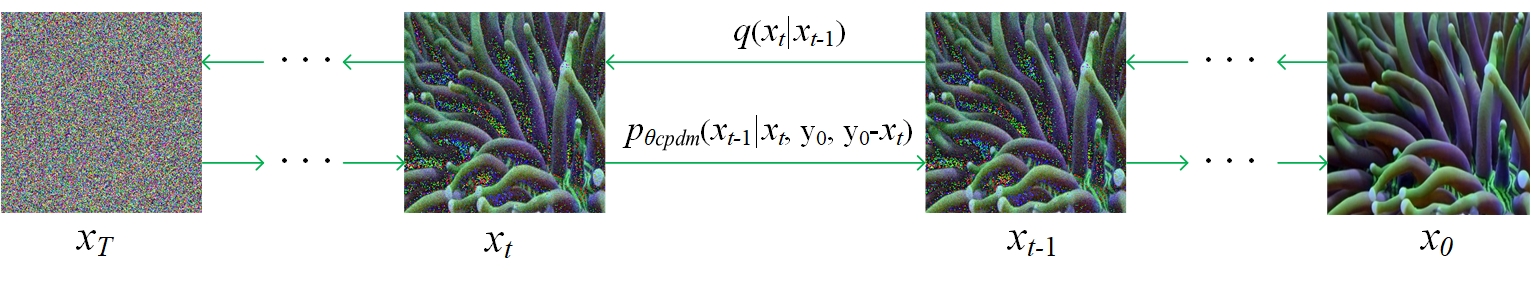}}

  % \centerline{(a) Result 1}\medskip
% \end{minipage}
\vspace{-1.0em}
\caption{Illustration of our conditional input module. The forward diffusion process denotes $q$ (from right to left), and the backward inference process denotes $p_{\theta_{cpdm}}$ (from left to right). $x_0$ and $y_0$ denote the clear and paired underwater images, respectively.}
\label{fig2}
% \vspace{-1.4em}
\end{figure*} 

\subsection{Diffusion Model for Image Generation}
As a generative model, the diffusion model has demonstrated impressive performance in numerous computer vision tasks. According to the presence or absence of conditions, the diffusion model can be categorized into conditional diffusion and unconditional diffusion \cite{croitoru2023diffusion}.

\noindent \textbf{Unconditional Diffusion.} Denoising Diffusion Probabilistic Model (DDPM) \cite{ho2020denoising, 2015Deep} draw inspiration from non-equilibrium thermodynamics \cite{Jarzynski1997Equilibrium} and are composed of two processes: forward noising process and backward denoising process. In the forward process, DDPM applies a Markov chain-based diffusion to gradually introduce noise into the original image until its distribution aligns with a standard Gaussian distribution. The backward process is the inverse of the forward process, where a sample is drawn from a standard Gaussian distribution, and the noise introduced during the forward process is gradually eliminated, resulting in the gradual generation of the target image. Denoising Diffusion Implicit Model (DDIM) \cite{song2020denoising} optimizes the sampling process in DDPM by transforming it into a non-Markovian process and enhancing sampling efficiency. The training process remains unchanged, while significant optimizations are made to the steps in the sampling process.

\noindent \textbf{Conditional Diffusion.} Conditional diffusion models are built upon the diffusion model and incorporate additional conditions, such as category, text, and image, to guide the diffusion and generation processes. Guided diffusion \cite{dhariwal2021diffusion} utilizes a classifier to classify the generated images, calculates gradients based on the cross-entropy loss between the classification score and the target category, and then employs these gradients to guide the next sampling. A notable advantage of this method is that it does not require retraining the diffusion model. Instead, guidance is added during the forward process to achieve the desired generation effect. Semantic guidance diffusion (SGD) \cite{liu2023more} introduces two forms of category guidance: reference graph-based guidance and text-based guidance. By designing corresponding gradient terms, the SGD method achieves specific guidance effects tailored to these different forms of category guidance. \cite{Li2023arXiv} applied the conditional diffusion model to unsupervised despeckling for AS-OCT images.

\section{Proposed method} 
\begin{figure*}[ht]
% \begin{minipage}[b]{1.0\linewidth}
  \centering
  \centerline{\includegraphics[width=18cm]{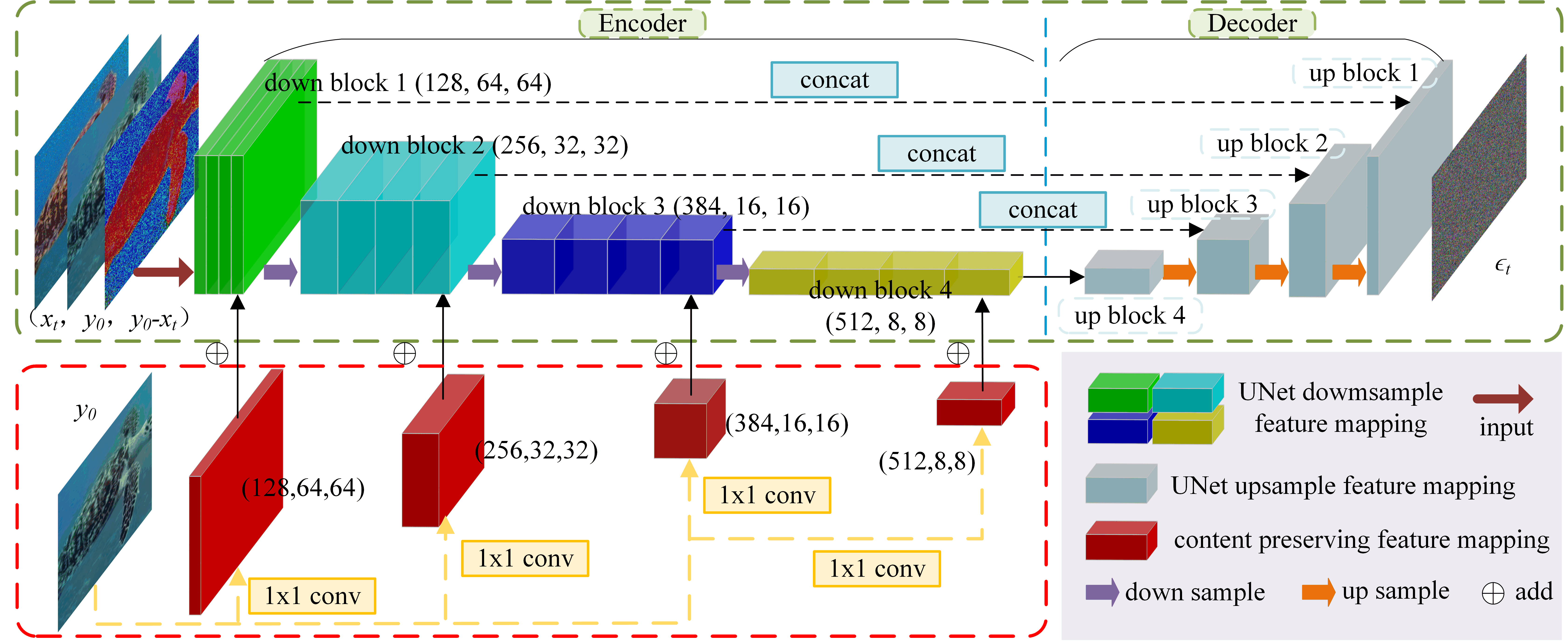}}

  % \centerline{(a) Result 1}\medskip
% \end{minipage}
 % \vspace{-0.5em}
\caption{Illustration of the proposed CPDM at time step $t$. Here, $y_0$ represents the to-be-enhanced underwater image, and $x_t$ denotes the noisy image of the current time step.}
\label{fig3}
 % \vspace{-1.4em}
\end{figure*}

This section provides a detailed description of the proposed Content-Preserving Diffusion Model (CPDM). Our CPDM includes the conditional input and content compensation modules, which are shown in Figures \ref{fig2} and \ref{fig3}. In the following, we will describe these two modules in detail. Note that the proposed CPDM is built upon the DDPM \cite{ho2020denoising}. To make the paper self-contained, we briefly introduce the mathematical background of the DDPM.

%The conditions play a crucial role in guiding the model to enhance specific features in the images. By adjusting the conditional inputs, the model can be guided to generate outputs that meet specific requirements or exhibit certain characteristics.  We incorporate specific conditions during the training of the noise prediction network.  To preserve the essential low-level information present in the original images, we introduce a content preservation module. This module is specifically designed to ensure that the enhanced images retain their fundamental structural and contextual details while improving overall image quality.
%\subsection{Denoising Diffusion Probabilistic Models}

The training of DDPM consists of a forward noising process and a backward denoising process. In the forward process, noise is step-by-step added to the original sample $x_0$ according to a progressively increasing diffusion rate $\beta_t$ ($\beta_t\in [0.0001, 0.02]$), resulting in the noisy image $x_t$ having a distribution that is closer and closer to the standard Gaussian distribution. 
% The conditional probability for the forward diffusion process can be written as
The forward diffusion process is defined as a Markov chain, which can be written as
\begin{align}
    \label{eq_q(x_{1:T}|x_0)}
    q(x_{1:T} | x_0) := \prod \limits_{t=1}^T q(x_t|x_{t-1}),\quad \quad \quad \quad \\
    q(x_t | x_{t-1}) := \mathcal{N}(x_t; \sqrt{(1-\beta_t)}x_{t-1}, \beta_t \textbf{I}), \quad
\end{align}
where $t\in [1, T]$ denotes the current time step of the diffusion process and $T$ represents the total number of diffusion steps. By defining $\alpha_t := 1 - \beta_t$ and $\bar{\alpha}_t := \prod_{i=1}^t \alpha_i$, the probability distribution $q(x_t|x_0)$ can be expressed as
\begin{equation}
    \label{eq_q_x_t|x_0}
    q(x_t|x_0) = \mathcal{N}(x_t;\sqrt{\bar{\alpha}_t}x_0, (1-\bar{\alpha}_t)\textbf{I}).
\end{equation}
The sampled value $x_t$ at time step $t$ for a given raw image $x_0$ can be expressed as
\begin{equation}
    \label{eq_x_t}
    x_t = \sqrt{\bar{\alpha}_t} x_0 + \sqrt{1-\bar{\alpha}_t} \epsilon,\quad \epsilon \sim \mathcal{N}(0, \textbf{I}).
\end{equation}
According to the computation of $x_t$, we can obtain that as $T$ becomes large, the value of $\sqrt{\bar{\alpha}_t}$ approaches $0$, and thus $x_t$ tends to $\epsilon$.

In the denoising process, solving the posterior distribution $p(x_{t-1}|x_t)$ is challenging. In practice, we can employ a neural network (denoted as $\theta$) to approximate this distribution, and the predicted distribution is denoted as $p_\theta(x_{t-1}|x_t)$. Assuming that the mean and variance of $p_\theta(x_{t-1}|x_t)$ are $\mu_\theta(x_t, t)$ and $\sigma_\theta(x_t, t)$, respectively, we can express $p_\theta(x_{t-1}|x_t)$ as
\begin{equation}
    \label{eq_p_theta(x_t-1|x_t)}
    p_\theta(x_{t-1}|x_t) = \mathcal{N}(x_{t-1}; \mu_\theta(x_t, t), \sigma_\theta(x_t, t)).  
\end{equation}
For the forward diffusion process, it can be inferred that given $x_0$ and $x_t$, the distribution of $p(x_{t-1}|x_t, x_0)$ can be expressed as
\begin{equation}
    \label{eq_p(x_t-1|x_t,x_0)}
    p(x_{t-1}|x_t, x_0) = \mathcal{N}(x_{t-1};\mu_t(x_t, x_0), \sigma_t),
\end{equation}
where $\mu_t(x_t, x_0) = \frac{\sqrt{\alpha}_t(1-\bar{\alpha}_{t-1})}{1-\bar{\alpha}_t} x_t + \frac{\sqrt{\bar{\alpha}_{t-1}}(1-\alpha_t)}{1-\bar{\alpha}_t} x_0$ and $\sigma_t = \frac{(1-\bar{\alpha}_{t-1})(1-\alpha_t)}{1-\bar{\alpha}_t}$.
By working out $x_0$ from Equation (\ref{eq_x_t}) and then substituting into $\mu_t(x_t, x_0)$, we can achieve
\begin{equation}
    \label{eq_mu_t(x_t, t)}
    \mu_t(x_t, t) = \frac{1}{\sqrt{\alpha}_t}(x_t - \frac{1-\alpha_t}{\sqrt{1-\bar{\alpha}_t}} \epsilon).
\end{equation}
Since the variance $\sigma_t$ of $p(x_{t-1}|x_t, x_0)$ is a constant, predicting $p(x_{t-1}|x_t, x_0)$ is equivalent to estimating $\mu_t(x_t, t)$. Thus, we may employ the network model $\theta$ to parameterize $\mu_\theta(x_t, t)$ by
\begin{equation}
    \label{eq_mu_theta(x_t, t)}
    \mu_\theta(x_t, t) = \frac{1}{\sqrt{\alpha}_t}(x_t - \frac{1-\alpha_t}{\sqrt{1-\bar{\alpha}_t}} \epsilon_\theta(x_t, t)),
\end{equation}
where $\epsilon_\theta(x_t, t)$ denotes the predicted value of added noise at time step $t$. During the training process, the goal is to make the predicted distribution $p_\theta(x_{t-1}|x_t)$ as close to the posterior $p(x_{t-1}|x_t, x_0)$ as possible. This is equivalent to minimizing $\mathbb{E}_{t,x_0,\epsilon}[||\mu_t(x_t, t) - \mu_\theta(x_t,t)||^2]$. An efficient and effective loss function for this minimization can be designed as
\begin{equation}
    \label{eq_loss}
    \mathcal{L}_{simple} = \mathbb{E}_{t,x_0,\epsilon}[||\epsilon-\epsilon_\theta(x_t,t)||^2].
\end{equation}

% where $\epsilon_\theta(x_t, t)$ represents the predicted value of the added noise at time step $t$. The inputs for $\epsilon_\theta(x_t, t)$ are $x_t$ and $t$. During the training process, the objective is to minimize the discrepancy between the predicted distribution $p_\theta(x_{t-1}|x_t)$ and the prior distribution $p(x_{t-1}|x_t, x_0)$. The loss function is defined as follows:

After the network model $\theta$ is well trained, predicting the noise introduced at time step $t$ becomes possible. The time step $t$ starts from $1$ and gradually increases until reaching $T$. As $T$ reaches a sufficiently large value, the variable $x_T$ at time step $T$ will follow a standard Gaussian distribution. During the sampling process, a pure noise (denoted as $n_T$) is sampled from the standard Gaussian distribution. Subsequently, the noise is input into the well-trained diffusion model $\theta$. Given the input of $n_t$ and time step $t$, the denoising $n_{t-1}$ in one step can be expressed as
\begin{equation}
    \label{eq_x_t-1}
    n_{t-1} = \frac{1}{\sqrt{\alpha}_t}(n_t - \frac{1-\alpha_t}{1-\sqrt{\bar{\alpha}_t}}(\epsilon_\theta(n_t, t))) + \sigma_t z,
\end{equation}
where $z\sim \mathcal{N}(0, \textbf{I})$. According to Equation (\ref{eq_x_t-1}), the noise caused by image degradation can be progressively removed from $n_T$ until a meaningful image ($n_0$) of the target domain is generated. With the basics of the DDPM, it is convenient to illustrate how to apply the DDPM to the UIE task with two new designs. In the following, we describe each of them in detail.

\subsection{Conditional Input Module}
Our CPDM for underwater image enhancement differs from the fundamental diffusion model in the training and sampling processes. First, we need a paired dataset $\{(x_0^i, y_0^i)\}, i=0,1,...,S$ for training, where $S$ is the size of the dataset, $y_0^i$ and $x_0^i$ denote the $i$-th raw degraded underwater image and its corresponding clear in-air image, respectively. For simplicity, we hereinafter use the sample ($x_0, y_0$) to represent an arbitrary training sample ($x_0^i, y_0^i$). In the training process, except for the input of the noisy image ($x_t$), we input the raw underwater image ($y_0$) at each time step $t$ as a supervisory condition, which can guide the diffusion model to generate the enhanced underwater images. Furthermore, we find that when applying to the UIE task, the diffusion model built upon the UNet \cite{ronneberger2015u} network structure is too simple to extract sufficient information from the raw image. To address this problem, an additional condition is introduced, which is the difference between the raw underwater image ($y_0$) and the noisy image ($x_t$) at the current time step $t$. By incorporating this difference $(y_0-x_t)$ as another conditional input, the network can extract more useful information and cues about $y_0$ and $x_t$. The extracted information helps the diffusion model to produce more accurate and visually appealing enhancements.

After introducing our conditional input ($y_0$ and $y_0-x_t$), the posterior probability of our diffusion model (denoted as $\theta_{cpdm}$) can be defined as 
\begin{equation}
\label{eq_p_theta(x_t-1|x_t, y_0)}
\begin{split}
    p_{\theta_{cpdm}}(x_{t-1}|x_t, &y_0, y_0-x_t)=\\
    &\mathcal{N}(x_{t-1}; \mu_{\theta_{cpdm}}(x_t, t, y_0, y_0-x_t),\\
    &\sigma_{\theta_{cpdm}}(x_t, t, y_0, y_0-x_t)). 
\end{split}
\end{equation}
Accordingly, the mean of the predicted noise can be written as 
\begin{equation}
\label{eq_mu_theta(x_t, t, y_0)}
\begin{split}
    & \mu_{\theta_{cpdm}}(x_t, t, y_0, y_0-x_t) =\\
    &\frac{1}{\sqrt{\alpha}_t}(x_t - \frac{1-\alpha_t}{\sqrt{1-\bar{\alpha}_t}} \epsilon_{\theta_{cpdm}}(x_t, t, y_0, y_0-x_t)).
\end{split}
\end{equation}
Therefore, our loss function $\mathcal{L}_{cpdm}$ can be defined as
\begin{equation}
    \label{eq_loss_y0}
     \mathcal{L}_{cpdm} = \mathbb{E}_{t,x_0,y_0,\epsilon}[||\epsilon-\epsilon_{\theta_{cpdm}}(x_t,t,y_0, y_0-x_t)||^2].
\end{equation}

%----------Fig3.-----------
%--------------------------

\subsection{Content Compensation Module}
In this part, we introduce a content compensation module to boost the ability of the information aggregation inside the noise prediction network. It is known that the UNet depends on a relatively straightforward network structure, and the UIE task demands the safeguarding of vital low-level features, including color, contour, edge, texture, and shape. On this basis, our content compensation module seamlessly integrates the low-level information extracted from the raw image $y_0$ into each layer of the UNet network. The training framework for time step $t$ is illustrated in Figure \ref{fig3}. As shown in  Figure \ref{fig3}, each layer of our UNet has four blocks, and the content compensation module inputs a low-level feature of the raw image into the last block of each layer in the encoder side (i.e., downsampling part). Such low-level features can always control the network to preserve the image content, which is beneficial to restore a high-quality target image corresponding to the input raw underwater image. The significance of the content compensation module lies in its ability to effectively retain the low-level features during the sampling process, thereby leading to an overall enhancement in the quality of the restored image. Algorithm \ref{alg:algorithm1} and Algorithm \ref{alg:algorithm2} illustrate our training and sampling processes, respectively.

\begin{algorithm}[th]
\caption{Training a denoising model $\epsilon_{\theta_{cpdm}}$}
\label{alg:algorithm1}
% \textbf{Parameter}: Optional list of parameters\\
% \textbf{Output}: Your algorithm's output
\begin{algorithmic}[1]
\STATE \textbf{Repeat} \\
\STATE  $(x_0, y_0) \sim q(x_0^i, y_0^i)$ \\
\STATE  $t \sim$ Uniform$(\{1, \ldots, T\})$ \\
\STATE  $\epsilon  \sim \mathcal{N}(0, \textbf{I})$ \\
\STATE  Take gradient descent step on \\
\quad $\nabla_\theta|| \epsilon - \epsilon_{\theta_{cpdm}}(x_0, t, y_0, y_0-x_t)||^2$\\
\STATE \textbf{Until} converged
\end{algorithmic}
\end{algorithm}

\begin{algorithm}[th]
\caption{Sampling for condition $y_0$}
\label{alg:algorithm2}
% \textbf{Parameter}: Optional list of parameters\\
% \textbf{Output}: Your algorithm's output
\begin{algorithmic}[1]
\STATE  \textbf{Sample} $x_T \sim \mathcal{N}(0, \textbf{I})$ and $y_0$ 
\FOR{$t=T, \ldots, 1$}
    \STATE $z \sim \mathcal{N}(\mathbf{0}, \mathbf{I})$ if $t > 1$, else $z = \mathbf{0}$
    \STATE $x_{t-1} = \frac{1}{\sqrt{\alpha}_t}(x_t - \frac{1-\alpha_t}{1-\sqrt{\bar{\alpha}_t}}(\epsilon_{\theta_{cpdm}}(x_t, t, y_0, y_0-x_t))) + \sigma_t z$
\ENDFOR
\STATE \textbf{Return} $x_0$
\end{algorithmic}
\end{algorithm}

%----------Fig4.-----------
%--------------------------
\begin{figure*}[thp]
% \begin{minipage}[b]{1.0\linewidth}
  \centering

  % \centerline{\includegraphics[height=9.5cm, width=17cm]{fig4-4.jpg}}
    % \begin{tikzpicture}
        {\includegraphics[height=9.5cm, width=17cm]{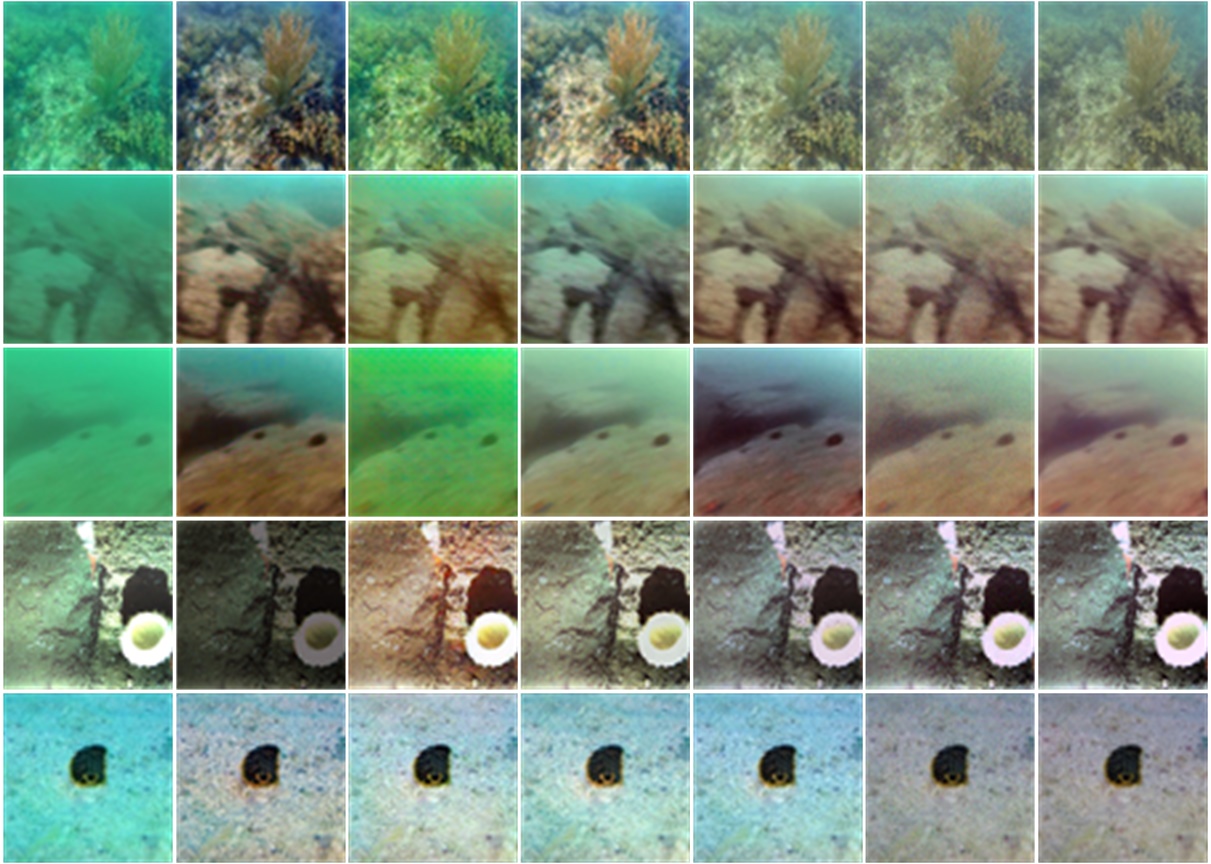}}
        % \node (text) at (image.south) [anchor=north, font=\footnotesize] 
        { Input \quad\quad\quad\quad\quad WaterNet \quad\quad\quad FUInE \quad\quad\quad\quad Ucolor \quad\quad\quad\quad Ushape \quad\quad\quad\quad CPDM \quad\quad\quad\quad\quad\quad GT}
      % \end{tikzpicture}
\vspace{-0.5em}
\caption{Visual comparison of enhanced underwater images by various methods on the Test\_L400 (LSUI) dataset. From left to right: original underwater image, results from WaterNet \cite{li2019underwater}, FUnIE \cite{9001231funie}, Ucolor \cite{9426457ucolor}, U-shape Transformer \cite{peng2023u}, our CPDM method, and the reference image.}
%  The highest PSNR after enhancement for each image is highlighted in green.
\label{fig4}
 % \vspace{-1.0em}
\end{figure*} 

%----------Fig5.-----------
%--------------------------
\begin{figure*}[h]
% \begin{minipage}[b]{1.0\linewidth}
  \centering
  % \centerline{\includegraphics[width=8.5cm]{fig1.png}}
  % \centerline{\includegraphics[height=9.5cm, width=17cm]{fig5-4.jpg}}
    % \begin{tikzpicture}
        {\includegraphics[height=9.5cm, width=17cm]{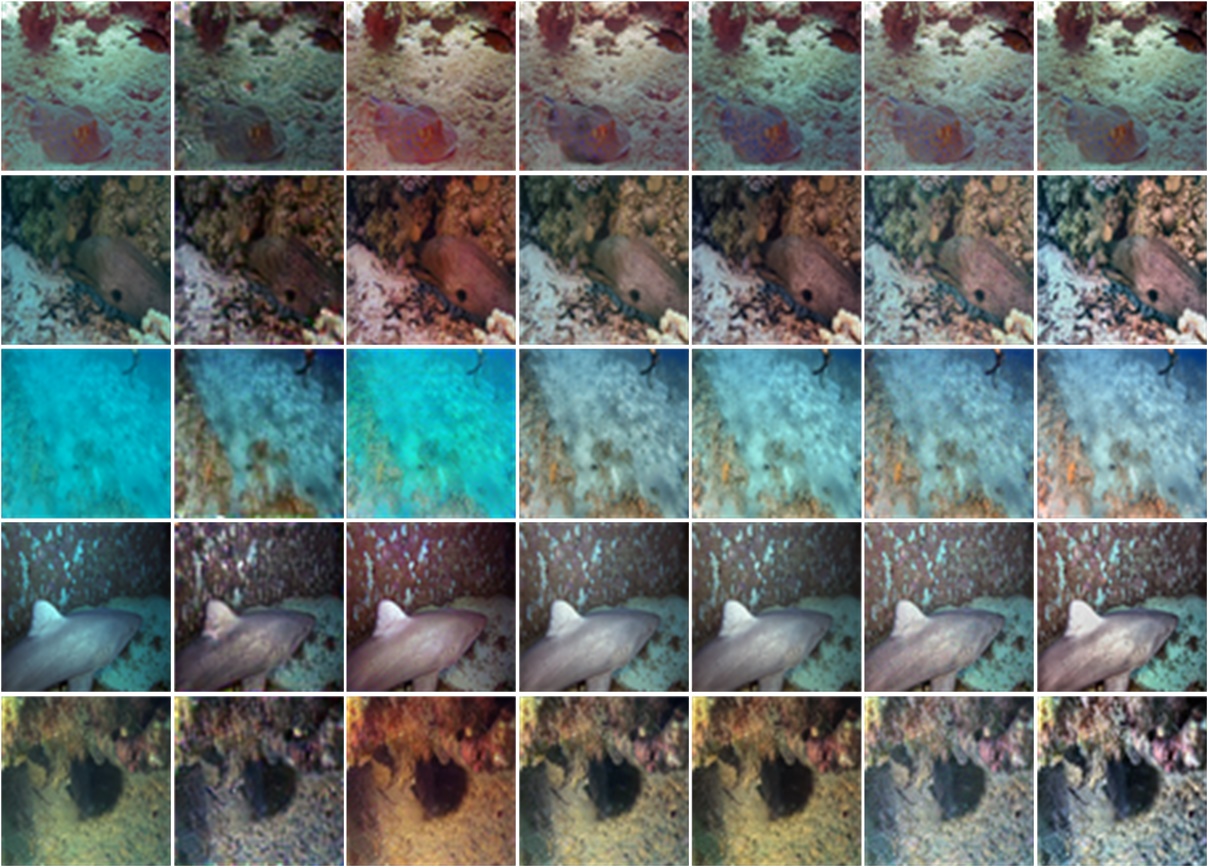}}
        % \node (text) at (image.south) [anchor=north, font=\footnotesize] 
        { Input \quad\quad\quad\quad\quad WaterNet \quad\quad\quad FUInE \quad\quad\quad\quad Ucolor \quad\quad\quad\quad Ushape \quad\quad\quad\quad CPDM \quad\quad\quad\quad\quad\quad GT}
      % \end{tikzpicture}
  % \centerline{(a) Result 1}\medskip
% \end{minipage}
 % \vspace{-1.0em}
\caption{Visual comparison of enhanced underwater images by various methods on the Test\_U90 (UIEB) dataset. From left to right: original underwater image, results from  WaterNet \cite{li2019underwater}, FUnIE \cite{9001231funie}, Ucolor \cite{9426457ucolor}, U-shape Transformer \cite{peng2023u}, our CPDM method, and the reference image.}
\label{fig5}
 % \vspace{-1.0em}
\end{figure*} 

% --------------table 1.----------------
% \usepackage{multirow}
\begin{table*}[th]
\large
% \Large
% \LARGE 
% \huge
% \Huge

% \setlength\tabcolsep{2pt}
% \vspace{-1.0em}
\begin{center}
\normalsize
\scalebox{0.93}{
\begin{tabular}{llllllllll}
\toprule[1pt]
\hline
% \textbf{\hline}
\multirow{2}{*}{Methods} & \multicolumn{3}{c}{Test\_L400} & \multicolumn{3}{c}{Test\_U90} & \multicolumn{3}{c}{Test\_E200} \\ \cline{2-10} 
& PSNR $\uparrow$      & SSIM $\uparrow$      &MSE $\downarrow$
& PSNR $\uparrow$      & SSIM $\uparrow$      &MSE $\downarrow$ 
& PSNR $\uparrow$      & SSIM $\uparrow$      &MSE $\downarrow$        \\ \hline
WaterNet \cite{li2019underwater}               
& 19.53       & 0.84    &0.0188    & 16.89          & 0.73     &0.0299 &17.44 & 0.74 &0.0253   \\ \hline
FUnIE  \cite{9001231funie}                   
& 23.46       & 0.86    &0.0115    & 18.46          & 0.74     &0.0276 &22.24 &0.88 &0.0086   \\ \hline
% UGAN  \cite{fabbri2018ugan}                   
% & 19.79       & 0.78    & $\times$  & 20.68        & 0.84     &$\times$  \\ \hline
% UIE-DAL  \cite{2019uie}                
% & 17.45       & 0.79    & $\times$  & 16.37        & 0.78     &$\times$  \\ \hline
Ucolor \cite{9426457ucolor}                  
& 21.77       & 0.88    &0.0092    & 20.59          & 0.83     &0.0132  &21.45 &0.86 &0.0092 \\ \hline
Restormer \cite{zamir2022restormer}                  
& 20.57       & 0.84    &0.0178    & 18.57          & 0.73     &0.0232 &19.08 &0.83 &0.0215  \\ \hline
Maxim \cite{tu2022maxim}                  
& 19.93       & 0.78    &0.0121    & 17.14          & 0.76     &0.0267 &18.05 &0.72 &0.0234  \\ \hline
U-shape \cite{peng2023u}                 
& 24.24       & \textbf{0.89}    &0.0065    & 20.07          & 0.78     &0.0137 &22.05 &0.86 &0.0089  \\ \hline
CPDM (Ours)                    
& \textbf{25.11}       & 0.88    &\textbf{0.0056}    & \textbf{21.07}          & \textbf{0.84}     &\textbf{0.0114}  &\textbf{23.24}  &\textbf{0.90} &\textbf{0.0062} \\ \hline
\bottomrule[1pt]
\end{tabular}}
\caption{
% Comparison with the top-performing methods on the LSUI dataset. The used metrics are PSNR, SSIM and MSE, and the best results are highlighted in bold.
Quantitative comparison of different UIE methods on the LSUI, UIEB, and EUVP datasets. The best results are highlighted in bold.
}
\label{table1}
\end{center}

\end{table*}

\section{Experiments}
\subsection{Dataset} Large Scale Underwater Image Dataset (LSUI) \cite{peng2023u} contains 4,279 image pairs. This dataset involves a rich range of underwater scenes (lighting conditions, water body types, and target categories) with good visual quality. We divide the LSUI into 3,879 pairs of training data (Train\_L) and 400 pairs of test data (Test\_L400). In addition, an Underwater Image Enhancement Benchmark (UIEB) dataset \cite{li2019underwater} is also used in our experiment. This dataset contains 890 data pairs, where 800 pairs are used for training, and the remaining 90 pairs are used for testing (Test\_U90). In addition to these two datasets, we select 200 pairs of underwater images from the Enhancing Underwater Visual Perception dataset (EUVP) \cite{9001231funie} for out-of-sample testing. This additional test dataset is referred to as Test\_E200. To facilitate the training of our diffusion model, we resize all image pairs into 64$\times$64 size. 

\subsection{Experimental Setting and Metrics} Our experiments are run on an RXT 3090 GPU. In the forward process, we set $T=1000$. Since the test dataset contains the reference images, we can compute the full-reference quality evaluation metrics such as PSNR \cite{6263880psnr}, SSIM \cite{5596999ssim}, and MSE \cite{li2019underwater}. These three metrics reflect the proximity to the reference, with higher PSNR values representing closer image content, higher SSIM values reflecting more similar structures and textures, and lower MSE values indicating smaller differences between the corresponding pixels of two images. Six mainstream UIE methods including WaterNet \cite{li2019underwater}, FUnIE \cite{9001231funie}, Ucolor \cite{9426457ucolor}, Restormer \cite{zamir2022restormer}, Maxim \cite{tu2022maxim}, and U-shape Transformer \cite{peng2023u} are selected for the performance comparison.

\subsection{Results} 
As shown in Table \ref{table1}, our Content-Preserving Diffusion Model (CPDM) method achieves promising results in quantitative metrics (PSNR, SSIM, and MSE). Specifically, on Test\_U90, our method outperforms all the compared methods in the three metrics. On Test\_L400, our method obtains 3.6\% improvement in PSNR, while exhibiting a bit slight decrease in SSIM compared to its best competitor. Furthermore, on Test\_E200, our CPDM method consistently outperforms all the competing techniques. Note that we conduct an extra test on the Test\_E200 dataset, without prior training on the EUVP dataset. This test can verify the generalization ability of the compared methods. As shown in Figures \ref{fig4} and \ref{fig5}, our CPDM method produces better visual results that are much closer to the reference images than its competitors. In particular, we can see from Figure \ref{fig4} that the enhanced images by our method preserve better color consistency with the reference ones compared to other methods. Regarding the luminance restoration, it can be observed from Figure \ref{fig5} that our CPDM achieves an obvious advantage over the compared methods.

The superior performance of our CPDM can be attributed to two key designs: conditional input module and content compensation module. Firstly, we introduce the conditional input during the training of the noise prediction network. By introducing the raw image and the difference between the raw image and the noisy image of the current time step as conditional input, the noise prediction model can iteratively refine useful characteristics from the conditional input. Through this step-by-step refinement, our method can restore high-quality target images. Secondly, the content compensation module plays a pivotal role in extracting the low-level features of the input images, as depicted in Figure \ref{fig3}. By integrating such features into the UNet network, the content compensation module ensures that the enhanced images possess the same content information as the raw images, such as edge, texture, and shape,  throughout the sampling process. This preservation of low-level features contributes significantly to the overall improvement in the quality of the enhanced images.

\begin{table}[th]
\large
% \Large
% \LARGE 
% \huge
% \Huge

% \vspace{-1.0em}
% \setlength\tabcolsep{4pt}
\begin{center}
\scalebox{0.7}{
\begin{tabular}{lllllll}
\toprule[1pt]
\hline
\multirow{2}{*}{Models} & \multicolumn{3}{c}{Test\_L400} & \multicolumn{3}{c}{Test\_U90} \\ \cline{2-7} 
& PSNR $\uparrow$      & SSIM $\uparrow$      &MSE $\downarrow$      
& PSNR $\uparrow$      & SSIM $\uparrow$      &MSE $\downarrow$         \\ \hline
model-A      &24.01       &0.87        &0.0074    &19.95      &0.80        &0.0164       \\ \hline
model-B      &24.72       &\textbf{0.88}        &0.0063    &20.31      &0.80        &0.0153       
\\ \hline
model-C      &24.26        &0.87        &0.0073    &21.03      &0.82        &0.0129
\\ \hline
model-D      & \textbf{25.11}       &\textbf{0.88}        &\textbf{0.0056}    &\textbf{21.07}      &\textbf{0.84}        &\textbf{0.0114}       \\ \hline
\bottomrule[1pt]
\end{tabular}}
\caption{Ablation study on the Test\_L400 and Test\_U90 datasets. Here, model-A represents only inputting $y_0$ as the input condition, model-B represents inputting both $y_0$ and $y_0$-$x_t$ as the input conditions, model-C represents inputting both $y_0$ and content compensation module as the input conditions, and model-D represents a full model.}
\label{table2}
\end{center}
% \vspace{-1.0em}

\end{table}

\subsection{Ablation Study} 
\begin{figure}[tb]
% \begin{minipage}[b]{1.0\linewidth}
  \centering
  % \centerline{\includegraphics[width=8.5cm]{fig1.png}}
  % \centerline{\includegraphics[width=8.5cm]{ablation4.pdf}}
     % \begin{tikzpicture}
        % \node (image) at (0,0) 
        {\includegraphics[width=8.5cm]{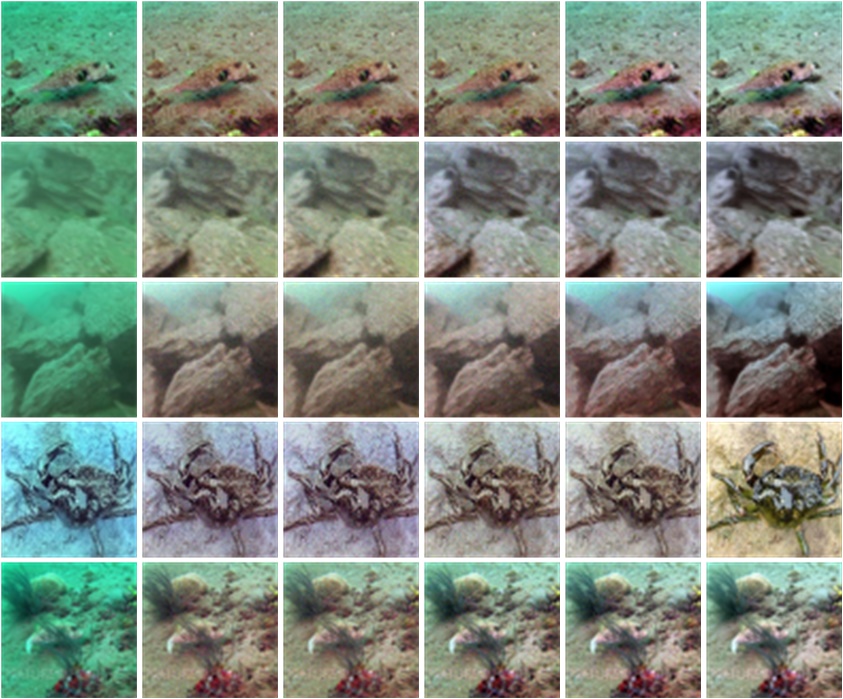}}
        % \node (text) at (image.south) [anchor=north, font=\footnotesize] 
        { Input  \hskip 0.5cm  model-A  \hskip 0.20cm   model-B \hskip 0.20cm   model-C \hskip 0.20cm    model-D  \hskip 0.30cm  GT \hskip 0.25cm}
      % \end{tikzpicture}
  % \centerline{(a) Result 1}\medskip
% \end{minipage}
    \vspace{-0.5em}
\caption{Comparison of visual effects on the Test\_L400 dataset in the ablation study.}
\label{fig6}
\vspace{-1.5em}
\end{figure}

To verify the effectiveness of each module designed in our method, we conduct a series of ablation experiments in a step-by-step addition manner. This way allows us to gradually assess the contribution of each module to the overall performance of CPDM. The numerical results of these ablation experiments are summarized in Table \ref{table2}, and the corresponding visualization effects are shown in Figure \ref{fig6}. 

\textbf{Base Model.} The visual effects show that utilizing only the raw image as the input condition (model-A) can produce semantically meaningful results. This showcases a certain potential of diffusion-based methods for enhancing underwater images. However, we can see from Table \ref{table2} that its numerical results arrive at the lowest level in all objective metrics. 

\textbf{Model with Conditional Input Module.} The performance of model-B is improved when the offset between the raw image and the noisy image of the current time step is added as the conditional input. Adding the dual-input condition significantly enhances the model's overall performance. Compared to model-A, incorporating the offset between the raw image and the noisy image of the current time step in model-B allows for further integration of the information of $x_t$ predicted at the current time step. As shown in Figure \ref{fig6}, the output images of model-B exhibit improved consistency in terms of color tone compared to those of model-A. Furthermore, as indicated by the quantitative metrics in Table \ref{table2}, model-B outperforms model-A in objective metrics. Our designed conditional input module incorporates the $y_0-x_t$ of the current time step as a conditional input into the diffusion model. Compared to the raw diffusion model, including a conditional input related to $x_t$ provides additional complementary information to the model. Thus, our designed conditional input module is proven to be beneficial for conditional diffusion tasks.

\textbf{Model with Content Compensation Module.} To validate the functionality of our designed content compensation module, we introduce model-C, which incorporates both $y_0$ and the content compensation module as inputs. Unlike model-B, model-C removes the input $y_0-x_t$ and includes a content compensation module instead. As shown in Figure \ref{fig6}, we can observe that all our models achieve good visual results, including model-A, model-B, and model-C. Interestingly, model-C exhibits improved color consistency closer to the real reference images than model-A and model-B. As presented in Table \ref{table2}, model-C outperforms models-A and model-B in terms of numerical results. The design of model-C incorporates the content compensation module on top of model-A. Our designed content compensation module can extract structural information from the input $y_0$ across different feature dimensions, which is then fed into different layers of the UNet architecture. This further enhances the encoding of input information and decoding of output in the UNet framework. Therefore, our designed content compensation module is also proven to be beneficial for conditional diffusion tasks.

\textbf{Full Model.} Finally, we can see that the full model (model-D) encompassing all modules achieves the best results. This outcome indicates that each module of the CPDM plays a specific role and contributes to its overall effectiveness. The step-by-step addition of these modules gradually enhances the model's capability, leading to improved image quality in the UIE task.
%----------Fig6.-----------
%--------------------------
\section{Conclusion}
In the article, we have attempted to adapt the diffusion model to the underwater image enhancement (UIE) task and have presented a Content-Preserving Diffusion Model (CPDM) for enhancing the quality of the restored underwater image. The proposed CPDM has demonstrated impressive performance compared with the leading UIE methods. We introduce two carefully designed conditional inputs, effectively guiding CPDM to generate high-quality results. Moreover, our designed content compensation module plays a crucial role throughout the training process, ensuring the content preservation of the raw image. Our CPDM works in an iterative refinement paradigm by embedding two modules into each time step in both the training and sampling processes, thereby preserving the image content in each denoising step and enhancing the quality of the restored images. Extensive experimental results validate the outstanding capabilities of CPDM in terms of numerical evaluations and visual effects. More importantly, the methodology designed in our CPDM can be easily extended to other conditional generative tasks.

\bibliography{sample}

\end{document}